\documentclass[journal]{IEEEtran}
\IEEEoverridecommandlockouts
\usepackage[T1]{fontenc}
\usepackage[cmex10]{amsmath}
\usepackage{amssymb,amsfonts}
\usepackage{mathtools}
\usepackage{physics}
\usepackage{breqn}
\usepackage{eqparbox}

\usepackage{tabularx}
\usepackage{graphicx}
\graphicspath{{./Imagens/}{./figures/}}
\DeclareGraphicsExtensions{.pdf,.jpeg,.png}
\usepackage{caption}
\usepackage{subcaption}
\usepackage{adjustbox}
\usepackage{float}
\usepackage{threeparttable}
\usepackage{rotating}

\usepackage{booktabs}     % For improved table lines
\usepackage{array}        % Better column formatting
\usepackage{multirow}
\usepackage{tabularx}
\usepackage{makecell}     % Thicker/thinner \hline
\usepackage{diagbox}

\usepackage{xcolor,soul,framed}
\colorlet{shadecolor}{yellow}

\usepackage[nocompress,space]{cite}

\usepackage{pifont}
\usepackage{bbding}
\usepackage{wasysym}
\usepackage{tikz}

\usepackage{mdwmath}
\usepackage{mdwtab}

\usepackage{textcomp}
\usepackage{url}
\usepackage{lipsum}
\usepackage{pdfrender}

\usepackage{hyperref}
\usepackage{cleveref}

\usepackage[
  separate-uncertainty = true,
  multi-part-units = repeat
]{siunitx}

\usepackage{orcidlink}

\usepackage[most]{tcolorbox}
\tcbuselibrary{skins,breakable}

\newtcolorbox{riskcodesbox}{
  breakable,
  enhanced,
  colback=green!4,            % fundo verde bem claro
  colframe=green!40!black,    % borda discreta
  boxrule=0.6pt,
  arc=2pt,
  left=6pt,
  right=6pt,
  top=6pt,
  bottom=6pt,
  coltitle=white,
  fonttitle=\bfseries,
  title=Risk Encoding and Numeric Taxonomy,
  attach boxed title to top left={
    xshift=0pt,
    yshift=-2mm
  },
  boxed title style={
    colback=green!50!black,   % barra superior verde escuro
    colframe=green!50!black,
    sharp corners,
    boxrule=0pt,
    left=6pt,
    right=6pt,
    top=2pt,
    bottom=2pt
  }
}

\definecolor{darkgreen}{rgb}{0.0, 0.5, 0.0}
\newif\ifshowmods
\showmodsfalse   % uncomment this line to create a clean doc

\ifshowmods

\else

\fi

 \makeatletter
\newcommand*\bigcdot{\mathpalette\bigcdot@{.5}}
\newcommand*\bigcdot@[2]{\mathbin{\vcenter{\hbox{\scalebox{#2}{$\m@th#1\bullet$}}}}}
\makeatother

\tcbset{
  enhanced,
  breakable,
  boxrule=0.45pt,
  arc=1pt,
  colback=gray!4,
  colframe=gray!55,
  coltext=black,
  coltitle=black,
  colbacktitle=gray!30,   % <<< ISSO muda o fundo do título
  left=5pt,right=5pt,top=5pt,bottom=5pt,
  fonttitle=\bfseries,
  width=\columnwidth,
}

\usepackage{xcolor}

\usepackage{algpseudocode}

\newenvironment{AlgoBox}[2]{%
\begin{tcolorbox}[title={Algorithm #1: #2}]
}{%
\end{tcolorbox}
}
\newtcolorbox{promptconstraintbox}{
  breakable,
  enhanced,
  colback=green!3,             % fundo ainda mais claro (regra/executor)
  colframe=green!35!black,
  boxrule=0.6pt,
  arc=2pt,
  left=6pt,
  right=6pt,
  top=6pt,
  bottom=6pt,
  coltitle=white,
  fonttitle=\bfseries,
  title=Prompt Constraints and JSON Output Enforcement,
  attach boxed title to top left={
    xshift=0pt,
    yshift=-2mm
  },
  boxed title style={
    colback=green!55!black,    % barra superior levemente mais forte
    colframe=green!55!black,
    sharp corners,
    boxrule=0pt,
    left=6pt,
    right=6pt,
    top=2pt,
    bottom=2pt
  }
}

\begin{document}

\title{

% LLM-Driven Human-Machine Interface for Smart Vehicle Digital Twins in the Smart Grid

% A Human-Centric Digital Twin for Smart Vehicles: Real-Time Environment Understanding via YOLO and GPT

% A Human-Centric Digital Twin for Smart Grid Vehicles Using YOLO and GPT

Dual-Stage LLM Framework for Scenario-Centric Semantic Interpretation in Driving Assistance

%An LLM-Based Framework for Scenario-Centric Semantic Interpretation in Driving Assistance

%Dual-Stage LLMs for Adaptive Driving Assistance and Sensor Calibration

%Dual-Stage Semantic Large Language Model Framework for Context-Adaptive Perception and Driving Assistance

%Dual-Stage Semantic LLM Framework for Context-Adaptive Perception and Driving Assistance

%Context-Aware Driving Assistance through Dual-Stage Large Language Models for Real-Time Scene Understanding and Sensor Calibration

}
 
% \author{ 
% Jean Douglas Carvalho\orcidlink{0009-0009-1372-7050}, Hugo Taciro Kenji \orcidlink{0009-0002-7958-4960},
% Ahmad~Mohammad~Saber\orcidlink{0000-0003-3115-2384},
% Glaucia Melo \orcidlink{0000-0003-0092-2171}, \\
% Max Mauro Dias Santos \orcidlink{0000-0001-7877-3554},
% and~Deepa~Kundur\orcidlink{0000-0001-5999-1847}
%         % <-this % stops a space
% % 
% % \thanks{ Ahmad Mohammad Saber and Deepa Kundur are with the
% % University of Toronto,
% % Toronto, ON,
% % Canada. } 
% }

% \author{ 
% Jean D. Carvalho\orcidlink{0009-0009-1372-7050}, Hugo T. Kenji\orcidlink{0009-0002-7958-4960},
% Ahmad~M.~Saber\orcidlink{0000-0003-3115-2384},
% Glaucia Melo\orcidlink{0000-0003-0092-2171},
% Max M. D. Santos\orcidlink{0000-0001-7877-3554}
% and~Deepa~Kundur\orcidlink{0000-0001-5999-1847}
%         % <-this % stops a space
% % 
% % \thanks{ Ahmad Mohammad Saber and Deepa Kundur are with the
% % University of Toronto,
% % Toronto, ON,
% % Canada. } 
% }

\author{ 
Jean Douglas Carvalho\orcidlink{0009-0009-1372-7050},
Hugo Taciro Kenji\orcidlink{0009-0002-7958-4960},
Ahmad~Mohammad~Saber\orcidlink{0000-0003-3115-2384}, \\
Glaucia Melo\orcidlink{0000-0003-0092-2171}, 
Max Mauro Dias Santos\orcidlink{0000-0001-7877-3554}, and~Deepa~Kundur\orcidlink{0000-0001-5999-1847}
}

\maketitle

\vspace{-5em}  % Adjust as needed (e.g., -2em)

\begin{abstract}

Advanced Driver Assistance Systems (ADAS) increasingly rely on learning-based perception, yet safety-relevant failures often arise without component malfunction, driven instead by partial observability and semantic ambiguity in how risk is interpreted and communicated. This paper presents a scenario-centric framework for reproducible auditing of LLM-based risk reasoning in urban driving contexts. Deterministic, temporally bounded scenario windows are constructed from multimodal driving data and evaluated under fixed prompt constraints and a closed numeric risk schema, ensuring structured and comparable outputs across models. Experiments on a curated near-people scenario set compare two text-only models and one multimodal model under identical inputs and prompts. Results reveal systematic inter-model divergence in severity assignment, high-risk escalation, evidence use, and causal attribution. Disagreement extends to the interpretation of vulnerable road user presence, indicating that variability often reflects intrinsic semantic indeterminacy rather than isolated model failure. These findings highlight the importance of scenario-centric auditing and explicit ambiguity management when integrating LLM-based reasoning into safety-aligned driver assistance systems.

\end{abstract}

\section{Introduction}

Advanced Driver Assistance Systems (ADAS) are increasingly deployed to enhance road safety by supporting drivers in perception, hazard anticipation, and decision-making. Despite substantial progress in sensing technologies and learning-based perception, the reliability of ADAS remains fundamentally constrained in real-world urban environments, where occlusions, illumination variability, and dense interactions with vulnerable road users (VRUs) are common~\cite{sun2024ensuring,feng2021multimodal,sezgin2023adverseweather}. In such settings, safety risks frequently arise not from component failures but from limitations in how complex traffic situations are perceived, interpreted, and communicated to drivers.

From a safety engineering standpoint, these limitations are not fully addressed by traditional Functional Safety frameworks, such as ISO~26262, which primarily address hazards arising from system malfunctions~\cite{iso26262}. Instead, many critical ADAS risks arise when systems operate as intended yet exhibit insufficient performance under ambiguous, uncertain, or unforeseen conditions. This class of fault-free hazardous behavior is explicitly addressed by the Safety of the Intended Functionality (SOTIF) framework standardized in ISO~21448, which highlights perception limitations, semantic ambiguity, and foreseeable misuse as primary safety concerns~\cite{iso21448}. Consequently, understanding and managing semantic uncertainty in perception and interpretation has become a central challenge for safety-oriented ADAS validation.

Addressing this challenge increasingly requires moving beyond raw sensor streams toward structured, scenario-centric representations. Scenario abstraction enables the decomposition of complex driving interactions into discrete, comparable, and auditable units, thereby supporting systematic and reproducible evaluation at scale. Frameworks such as MetaScenario formalize this concept by organizing multimodal driving data into semantically indexed scenario representations, facilitating consistent comparison across heterogeneous conditions while preserving real-world complexity~\cite{chang2023metascenario}. Importantly, such representations focus on scenario description and organization, remaining largely agnostic to how higher-level reasoning components interpret risk and intent within those scenarios.

Within this scenario-centric paradigm, recent advances in Large Language Models (LLMs), including multimodal variants, introduce new opportunities for high-level semantic interpretation of driving situations. Unlike conventional perception models that operate at the signal or object-detection level, LLMs can integrate heterogeneous contextual cues, such as object relations, ego-vehicle dynamics, infrastructure attributes, and environmental context, into structured, human-interpretable assessments of traffic situations~\cite{chen2023driving_llm,ge2024llm_os,gridllm}. In this work, LLMs are not treated as perception, control, or decision-making modules, but as interpretative and analytical components capable of reasoning over scenario-bounded contextual evidence~\cite{llmpowered,ge2024llm_os}.

However, introducing LLMs into safety-critical driving contexts raises fundamental questions regarding reliability, consistency, and safety alignment. Different models may produce divergent semantic interpretations when exposed to identical traffic scenarios, particularly under partial or ambiguous perception, leading to variability in risk attribution and VRU assessment~\cite{ieee_llm_comparison,ammar2025survey}. Rather than representing isolated model errors, such divergences may reflect intrinsic semantic ambiguity in the available evidence. Without a structured and reproducible evaluation framework, these differences remain difficult to quantify, compare, or audit in a way that supports safety-oriented development.

Motivated by these challenges, this work investigates how LLM-based semantic reasoning behaves when exposed to identical real-world driving scenarios under deterministic, scenario-centric representations and constrained semantic outputs. By grounding the analysis in reproducible, scalable scenario abstractions derived from multimodal driving data, the proposed framework enables systematic, large-scale comparison of inter-model differences in risk interpretation, evidence use, and sensitivity to vulnerable road users. In doing so, this study positions LLMs as interpretable cognitive probes for ADAS evaluation, providing a controlled methodology to expose semantic ambiguity and assess its implications for safety-aligned system design.

The remainder of this paper is organized as follows: Section \ref{background} establishes the technical and conceptual foundations of the work, characterizing limitations in conventional ADAS pipelines and formalizes the problem of context-dependent perception. Section \ref{datasetSection} describes the multimodal data platform and the preprocessing pipeline used to generate the normalized 1 Hz semantic substrate. Section \ref{architectureSec} details the system architecture, focusing on the deterministic scenario construction and the prompted LLM evaluation framework. Section \ref{experimentSec} presents the experimental setup, including the collection protocol for safety-critical "near-people" scenarios and the language models evaluated. Section \ref{resultsSec} provides a quantitative and qualitative analysis of the experimental results, followed by the concluding remarks in Section \ref{conclusion}.

% A comparative evaluation 

% \begin{itemize}
% \vspace{-0.3em}\item \Ablue{Introduction of} a modular digital twin framework that combines visual perception, vehicle telemetry, and LLM reasoning; and Introduction of a structured prompting system that translates sensor data into interpretable alerts;
% % \item \Ablue{Introduction of} a structured prompting system that translates sensor data into interpretable alerts;
% \item \Ablue{To validate the above contributions, the proposed framework is comparatively evaluated} 
% % A comparative evaluation 
% across real-world driving scenarios, demonstrating risk awareness and alignment with human expectations.
% \end{itemize}

% In this \Ablue{regard}, the main contributions of this work include:
% \begin{itemize}
% \item A modular digital twin framework that combines visual perception, vehicle telemetry, and LLM reasoning;
% \item A structured prompting system that translates sensor data into interpretable alerts;
% \item A comparative evaluation across real-world driving scenarios, demonstrating risk awareness and alignment with human expectations.
% \end{itemize}

\section{Background and Problem Statement} \label{background}

This section outlines the technical foundations of the proposed framework and motivates scenario-centric auditing of semantic risk interpretation in ADAS.

ADAS rely fundamentally on accurate perception and timely interpretation of complex traffic scenes to support safe driving decisions~\cite{zhu2017overview_perception,neumann2024adas_analysis}. Modern perception pipelines predominantly employ camera- and radar-based sensing combined with computer vision and signal processing algorithms to detect objects, estimate motion, and infer driving context~\cite{sun2020mimo_radar_adas,yeong2025multisensor_xai_survey}. Despite significant progress in deep learning--based perception, current ADAS architectures remain largely reactive and context-agnostic under rapidly changing environmental conditions such as low illumination, adverse weather, occlusions, or dense urban traffic~\cite{feng2021multimodal,sun2024ensuring,sezgin2023adverseweather}.

Beyond raw perception performance, a core practical limitation is that many ADAS deployments provide limited transparency regarding how warnings are triggered and how risk is framed in ambiguous situations. This becomes particularly problematic in mixed traffic and partial observability conditions, where human drivers expect context-aware and interpretable feedback~\cite{omeiza_explanations_survey,explainable_ai_trust_ad}. As a result, the safety relevance of an ADAS output is not solely determined by whether objects are detected, but also by how risk is semantically interpreted and communicated under uncertainty.

\subsection{Limitations of Conventional ADAS Pipelines}

Most existing ADAS pipelines decouple perception, sensor control, and decision-making into independent modules~\cite{velasco2020autonomous_arch}. Vision models such as YOLO or semantic segmentation networks operate on raw sensor inputs with limited awareness of sensor configuration or environmental semantics~\cite{feng2021multimodal,sun2024ensuring,lin2025domain_adaptation_adverse_weather}. Sensor calibration, when present, is typically performed offline or through predefined rule-based logic, which constrains adaptability and generalization across diverse conditions.

From an engineering perspective, this modular decoupling yields two persistent gaps. First, downstream risk reasoning (whether implemented as rules, heuristics, or learned modules) may inherit perceptual uncertainty without a systematic mechanism to expose how ambiguity propagates into safety-relevant interpretations. Second, explainability often remains limited: the system may trigger warnings or interventions without conveying the underlying rationale in a form that supports user trust and post hoc analysis~\cite{omeiza_explanations_survey,explainable_ai_trust_ad}.

Formally, conventional approaches optimize perception models for detection accuracy under fixed assumptions:

\begin{equation}
\scalebox{1}{
$\min_{\boldsymbol{\phi}} \; \mathbb{E} \left[ \mathcal{L}\big(f_{\boldsymbol{\phi}}(\mathbf{y}_t), \mathbf{x}_t \big) \right]$
}
\end{equation}

\noindent where $f_{\boldsymbol{\phi}}$ denotes a perception model with parameters $\boldsymbol{\phi}$. Notably, this formulation treats sensor behavior as exogenous, omitting adaptive control over $\boldsymbol{\theta}_s$. While sensor adaptation and perception robustness are active research topics, they are orthogonal to the present work: this paper does not propose sensor adaptation or vehicle control mechanisms. Instead, it focuses on the subsequent layer, how higher-level semantic risk interpretation can diverge even when the input evidence is held fixed.

To address structural limitations in scenario handling, scenario-centric representations have been proposed in the literature, decomposing complex driving situations into discrete, comparable, and auditable units. Frameworks such as MetaScenario~\cite{chang2023metascenario} formalize scenarios as first-class entities, enabling systematic description, indexing, and retrieval of driving situations across datasets and experiments. While such representations significantly improve reproducibility and comparability at the scenario level, they remain largely agnostic to how higher-level reasoning components interpret risk, intent, and causality within those scenarios. Consequently, the use of scenario-centric abstractions to explicitly expose semantic ambiguity and inter-model divergence in risk interpretation remains an open problem.

\subsection{Perception as a Context-Dependent Estimation Problem}

From a systems perspective, perception in ADAS can be modeled as a state estimation problem in which the vehicle infers a latent scene state $\mathbf{x}_t$ from noisy sensor observations $\mathbf{y}_t$~\cite{aeberhard2012track_to_track_imf}:

\begin{equation}
\scalebox{1}{
$\mathbf{y}_t = h(\mathbf{x}_t, \boldsymbol{\theta}_s) + \boldsymbol{\nu}_t$
}
\end{equation}

\noindent where $h(\cdot)$ represents the sensor observation model, $\boldsymbol{\theta}_s$ denotes sensor configuration parameters (e.g., camera exposure, ISO, shutter speed, radar range), and $\boldsymbol{\nu}_t$ captures measurement noise. In practice, $\boldsymbol{\theta}_s$ is commonly fixed or adjusted via offline calibration and handcrafted heuristics, implicitly assuming quasi-stationary environmental conditions~\cite{lin2025domain_adaptation_adverse_weather}.

Real-world driving violates this assumption. Environmental factors such as illumination, weather, road geometry, and traffic density dynamically alter the effective observation process, thereby degrading detection confidence and segmentation fidelity when parameters remain static. This degradation propagates downstream, increasing the likelihood of missed detections, false alerts, or semantically ambiguous evidence. Importantly, even when perception outputs are available, their interpretation at the level of risk, intent, and hazard type may remain underdetermined, a property that becomes critical when assessing safety behavior in complex urban scenes~\cite{feng2021multimodal,sun2024ensuring,sezgin2023adverseweather}.

In this work, these perception-related uncertainties are treated as upstream conditions that motivate controlled analysis. The goal is not to optimize $h(\cdot)$ or adapt $\boldsymbol{\theta}_s$, but to study how fixed, scenario-bounded evidence can nonetheless yield divergent high-level risk interpretations.

\subsection{Emergence of Semantic Reasoning and Language Models}

Recent advances in Large Language Models (LLMs), particularly multimodal variants, introduce a new capability: semantic reasoning over heterogeneous evidence streams~\cite{chen2023driving_llm,ge2024llm_os}. Unlike traditional perception models that operate primarily in feature space, LLM-based components can integrate visual cues, vehicle telemetry, and contextual knowledge to generate structured interpretations of scene semantics, risk factors, and intent~\cite{gridllm,llmpowered}.

Prior work has explored LLMs for post hoc interpretation, human--machine interfaces, and offline analysis in autonomous driving contexts~\cite{chen2023driving_llm,ge2024llm_os}. Although the broader literature has also discussed leveraging high-level semantic representations for adaptive sensing or orchestration, such control-oriented roles are explicitly outside the scope of this paper. Here, LLMs are treated strictly as interpretative and analytical components: given identical, scenario-bounded contextual evidence, the objective is to characterize how different models frame risk, select evidence, and attribute causes in the face of semantic ambiguity.

This framing emphasizes a distinct validation challenge: if LLM-based interpretative modules are to support safety arguments, their behavior must be consistent enough to be audited, or at a minimum, their divergences must be measurable and explainable under controlled conditions.

\subsection{Safety, SOTIF, and Cybersecurity Foundations for LLM-Enabled ADAS}

The integration of LLM-based components into ADAS must be framed within established automotive safety and security standards. In safety-critical vehicular systems, nominal correctness is insufficient; behavior must be demonstrably safe, predictable, and robust under both fault conditions and performance limitations. Accordingly, the analysis of LLM-augmented ADAS architectures is commonly discussed through the combined lenses of Functional Safety (ISO 26262)~\cite{iso26262}, Safety of the Intended Functionality, SOTIF (ISO 21448)~\cite{iso21448}, and Cybersecurity (ISO 21434)~\cite{iso21434}.

Rather than asserting that these standards ``fail'' to address perception limitations, a more precise interpretation is that they acknowledge uncertainty and performance limitations (particularly under SOTIF), while leaving open the question of how semantic ambiguity should be exposed, compared, and audited in concrete engineering workflows. In this sense, the present work is complementary: it neither proposes a certified safety solution nor modifies the standards. Instead, it provides a controlled analytical methodology that makes ambiguity and interpretative divergence explicit, thereby supporting more structured safety arguments when LLM-based interpretative modules are considered.

\subsection{Problem Statement and Research Gap}

The considerations above reveal a critical gap in current ADAS research and validation practices. While perception pipelines and emerging semantic reasoning components can produce rich interpretations of driving scenes, there is no established framework that systematically exposes, compares, and audits these interpretations under identical real-world conditions. In particular, without structured, scenario-centric representations, it becomes difficult to isolate how semantic ambiguity in upstream evidence propagates into higher-level risk reasoning, and whether observed disagreements reflect intrinsic ambiguity or model-specific bias.

This paper addresses this gap by adopting scenario-based semantic abstraction as the fundamental unit of analysis. Driving situations are represented as deterministic, temporally bounded scenarios that can be reproduced, queried, and evaluated consistently across experiments. This design enables systematic, scalable evaluation, allowing large collections of identical traffic scenarios to be analyzed under controlled conditions without confounding variation in input evidence.

Within this framework, identical scenarios are presented to different LLMs under fixed prompts and bounded contextual inputs. Rather than treating disagreement as an isolated model failure, divergences in semantic interpretation are explicitly captured as measurable analytical signals. The following sections operationalize this approach through a multimodal data platform, a deterministic scenario-construction pipeline, and an evaluation protocol designed to support reproducible, large-scale inter-model comparison.

% \section{Overview of The Proposed Digital Twin Framework}

% ======================================================
% ======================================================
\section{Dataset Description and Pre-Processing Pipeline} \label{datasetSection}

The effectiveness of the proposed framework relies on transforming raw, high-frequency sensor data into structured, scenario-centric representations that can be audited for semantic consistency. To bridge the gap between low-level signal processing and high-level cognitive reasoning, we have developed a data-centric architecture that normalizes and aligns heterogeneous input streams. This section details the multi-layered pipeline for consolidating these sources into a unified temporal substrate, ensuring that subsequent scenario-based evaluations remain reproducible across diverse driving environments and model configurations.

\subsection{Multimodal Data Platform}
\label{sec:data_inputs}

\begin{figure}[b]
    \centering
    \includegraphics[width=1\columnwidth]{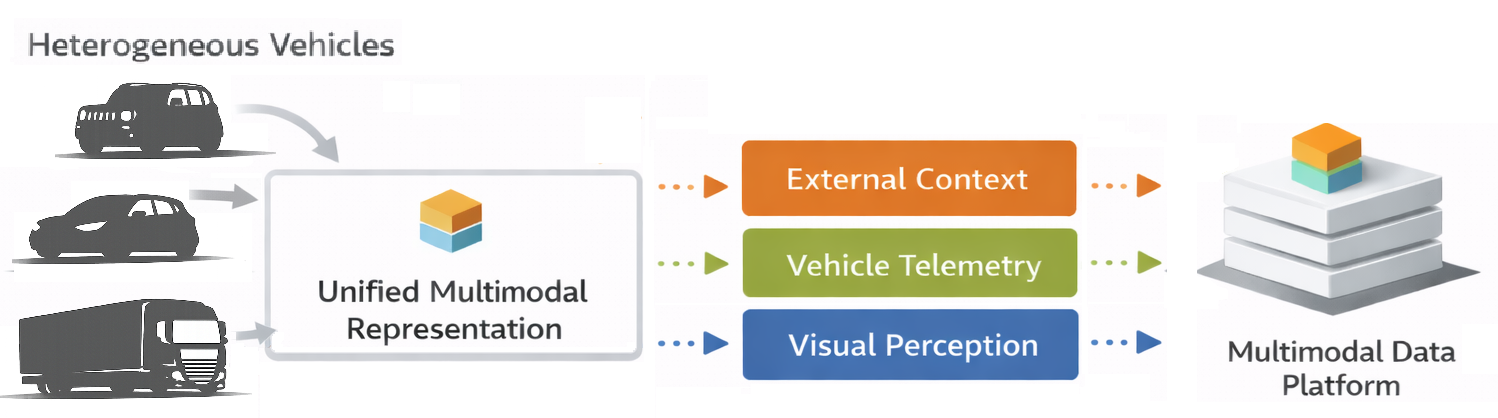}
    \caption{\textbf{Unified Multimodal Dataset: }Data from heterogeneous vehicles are abstracted into a unified multimodal representation that jointly encodes visual perception, vehicle telemetry, and external contextual information within the multimodal data platform.}

    \label{fig:input}
\end{figure}

The platform is built upon a proprietary multi-modal dataset collected from instrumented vehicles operating in real urban environments, available at \href{https://carcara.onrender.com/#/}{carcara.com}. The dataset integrates visual perception, vehicle telemetry, and external contextual information, and deliberately adopts a sensing configuration that excludes LiDAR sensors. This architectural choice reflects a trade-off that prioritizes reduced instrumentation cost, simplified deployment, and high replicability across heterogeneous vehicles and urban contexts, while remaining sufficient to support semantic scene understanding and risk-oriented interpretation tasks. Lowering the marginal cost and complexity of sensing is widely recognized as a key factor in enabling the collection of representative real-world driving data at scale, particularly when data acquisition spans multiple vehicles and long operational periods.

The primary contribution lies in the architectural organization of heterogeneous data into a scenario-centric semantic abstraction that is independent of the specific sensing configuration. In this sense, the proposed pipeline is not tied to a particular dataset or sensor suite; rather, it defines a generalizable representation strategy that can be applied to any multimodal driving dataset that provides synchronized perception, vehicle state, and contextual information.

Visual modalities operate at higher acquisition rates and provide the primary source of perceptual grounding. Vehicle-mounted cameras capture video at approximately 15~fps. These visual streams are processed to extract object detections, semantic scene segmentation, and lane geometry estimates. Although computed at video frequency, these outputs are not retained as raw frame-level data, but as compact semantic descriptors suitable for multimodal alignment.

Despite heterogeneity in sampling rates across modalities, all data streams are ultimately aligned to a common temporal resolution of 1 Hz. This temporal alignment constitutes a deliberate abstraction step that enables consistent multimodal fusion and subsequent scenario-based reasoning, while avoiding unnecessary dependence on high-frequency raw signals. Rather than treating the dataset as a collection of independent sensor logs, the platform organizes all available information into a temporally structured semantic substrate, explicitly designed to support scenario-level interpretation and reproducible evaluation. Figure~\ref{fig:input} summarizes this abstraction process, illustrating how heterogeneous multimodal inputs are consolidated into a unified 1~Hz semantic representation.

\subsection{Processing and Fusion}
\label{sec:processing_pipeline}

\subsubsection{Ingestion and Offline Heavy Processing}
\label{sec:ingestion_processing}

Visual perception tasks include semantic scene segmentation, lane geometry extraction, and object detection, augmented by a tracking process that associates successive detections over time. This tracking stage introduces temporal continuity into the perceptual representation by establishing persistent object identities across frames, thereby enabling the construction of dynamic entities whose evolution can subsequently be reasoned about at the semantic level. All visual processing stages are executed entirely offline, as they operate at video frequency, incur high computational cost, and are therefore intentionally decoupled from any real-time or online reasoning loop.

Let $\mathcal{D}_t=\{d_{t}^{1},\dots,d_{t}^{N_t}\}$ be the set of detections at time $t$. 
Tracking defines an association function
\[
a_t:\mathcal{D}_t \rightarrow \mathcal{I}\cup\{\varnothing\},
\]
which assigns each detection to a persistent identity $i\in\mathcal{I}$ (or $\varnothing$ for unmatched detections), thereby inducing temporal continuity across frames.

\subsubsection{Internal Fusion}
\label{sec:multimodal_fusion}

Following modality-specific processing, the platform performs multimodal fusion to integrate complementary information sources into a unified semantic representation of the driving scene. This fusion stage operates deterministically, relying on temporal alignment and spatial consistency across modalities rather than on learned or heuristic decision logic, and is designed to consolidate heterogeneous observations into a coherent semantic structure that can serve as a stable substrate for subsequent scenario abstraction and higher-level interpretation. The fusion of object detection and tracking outputs yields persistent semantic entities, each associated with spatial attributes, temporal extent, and categorical information, and explicitly linked to lane geometry estimates. This association enables the derivation of spatial relationships such as lane affiliation, relative position with respect to the ego vehicle, and proximity to road boundaries, transforming isolated detections into context-aware scene elements that retain temporal continuity and spatial meaning.

\subsubsection{External contextual enrichment}
\label{sec:multimodal_fusion}

In addition to dynamic perceptual inputs, the scene representation is enriched with external contextual information obtained through external APIs. Static and semi-static road attributes, including road type, number of lanes, and sidewalk presence, are integrated from digital map services such as OpenStreetMap, while basic environmental descriptors, including weather conditions, are retrieved from external weather APIs and provide complementary situational context. Importantly, the multimodal fusion stage does not perform semantic reasoning, risk assessment, or interpretive judgment; its role is strictly limited to organizing and aligning heterogeneous data into a consistent semantic substrate, on which scenario abstraction and cognitive evaluation can subsequently be conducted in a controlled, reproducible, and safety-oriented manner.

\subsection{Data Layer, Services and Platform Consumption}
\label{sec:data_layer_services}

\subsubsection{Normalized Data Layer}
\label{sec:normalized_data_layer}

Following multimodal fusion and temporal alignment, all scene representations are consolidated into a single unified data layer organized at a fixed temporal resolution of 1~Hz. Each temporal unit within the unified layer corresponds to a semantic snapshot of the driving scene. These snapshots aggregate ego-vehicle dynamics, persistent object representations, road and infrastructure attributes, environmental context, and acquisition-level metadata into a coherent structure that can be queried and recomposed deterministically. By concentrating all relevant modalities into a single temporally indexed representation, the platform enables complete scene reconstruction and interpretation from a single query, supporting both interactive exploration and automated analysis workflows. The explicit separation between raw data, processed perceptual outputs, and the normalized semantic layer ensures that computationally expensive perception stages are not repeated, while preserving flexibility for complex querying and scenario recomposition. In this organization, indexing is treated as a semantic and temporal construct rather than as a database-specific mechanism, enabling efficient access to context elements and consistent reuse of identical scene representations across multiple evaluation campaigns. This layered design supports scalability while preserving the methodological rigor required for reproducible scenario-based analysis.

\subsubsection{Backend Data Services}
\label{sec:backend_services}

Built directly on top of the normalized data layer, the platform backend provides online data services that expose multimodal scene representations in a controlled, efficient, and reproducible manner. Rather than coupling consumers to implementation-specific storage details, the service layer can be formalized as a deterministic query operator over the unified temporal representation. The service layer supports context-oriented querying, enabling retrieval of complete temporal scenes filtered by semantic, spatial, and dynamic criteria, including road characteristics, object configurations, environmental conditions, and ego-vehicle behavior. In addition to fine-grained filtering, the backend ensures consistent retrieval and organization of temporal contexts, enabling identical scenario definitions to be accessed, reused, and compared across different analytical processes.

Let $\mathcal{U}$ denote the normalized scene state at a fixed temporal resolution of 1~Hz, indexed by acquisition identifier $a$ and discrete time $t$. Let $\theta$ denote a structured scenario query encoding the semantic, spatial, behavioral, and temporal constraints that define a scenario of interest. The backend executes this query over the normalized representation and materializes the corresponding scenario context $\mathcal{S}$, which is a structured collection of normalized scene states indexed by the acquisition-time pairs selected by the query.

This operation can be expressed as
\begin{equation}
\mathcal{S} \;=\; \mathcal{B}\!\left(\mathcal{U}(a,t),\, \theta\right),
\label{eq:backend_query}
\end{equation}
where $\mathcal{B}(\cdot)$ denotes the deterministic backend query operator responsible for scenario construction. The set of valid $(a,t)$ indices defining the temporal extent of $\mathcal{S}$ is implicitly induced by the query specification $\theta$, ensuring that the scenario $\mathcal{S}$ materializes all database time instants whose normalized scene states satisfy the query constraints, such as specific vehicle types, illumination conditions, weather context, road environment, or traffic density, while preserving their original temporal ordering.

By operating on the normalized 1~Hz representation, the service layer leverages a temporally sparse yet semantically dense abstraction that is significantly lighter than the original high-frequency sensor streams. This design creates the computational margin required for efficient online querying, scenario reuse, and scalable consumption across multiple platform interfaces.

\subsubsection{Consumption Interfaces}
\label{sec:platform_consumption}

\begin{figure}[t]
    \centering
    \includegraphics[width=0.55\columnwidth]{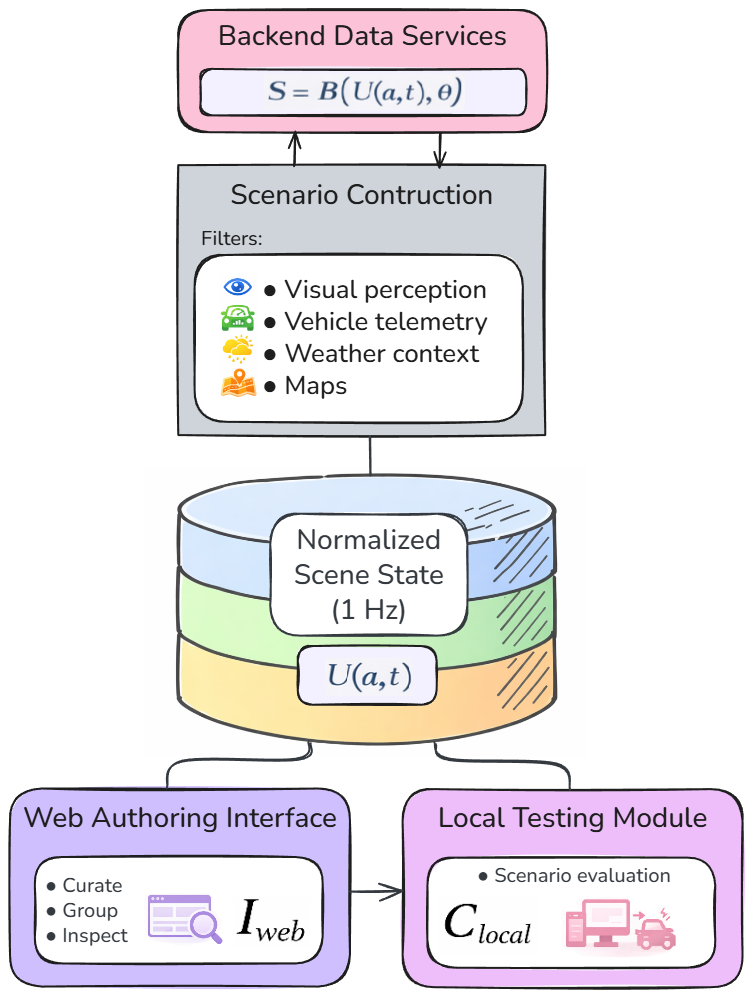}
    \caption{Deterministic scenario construction and consumption workflow. Normalized scene states $\mathcal{U}(a,t)$, indexed at 1~Hz, are queried by backend data services to materialize scenario snapshots $\mathcal{S}(a,t)$ through the deterministic operator $\mathcal{B}(\cdot)$ under a structured query specification $\theta$. Scenario construction combines semantic filters spanning visual perception, vehicle telemetry, weather context, and map information. The resulting scenarios are exposed consistently to both the web-based authoring interface for selection and inspection and the local testing module for scenario evaluation.}

    \label{fig:services}
\end{figure}

The backend data and services are accessed via complementary platform interfaces that support both interactive and automated workflows. The web-based interface provides a user-oriented environment for visual exploration of normalized scene representations, enabling synchronized inspection of perceptual outputs, contextual attributes, and temporal information for each acquisition. Within this interface, users can execute context-oriented queries to retrieve complete temporal scenes and curate selected temporal units into persistent scenario groupings, which explicitly reflect analytical interests and evaluation objectives while preserving the deterministic semantics of the underlying representation. In parallel, a local consumption interface supports external analytical modules that operate outside the web platform. This interface enables the systematic consumption of predefined scenario groupings, exposing structured multimodal inputs to automated workflows such as model-driven analyses or language-model-based evaluation pipelines.

Formally, the web interface materializes the scenario context as a canonical artifact
$\mathcal{S}$, which serves as the unified representation of a curated driving
scenario. This artifact is subsequently consumed by a local processing pipeline
through a dedicated consumption operator, without altering its semantic or temporal
structure. The relationship between interface, scenario, and consumption can be
expressed as
\begin{equation}
\mathcal{I}_{\text{web}} \;\rightarrow\; \mathcal{S} \;\xrightarrow{\;\mathcal{C}_{\text{local}}\;}\; \mathcal{O},
\label{eq:scenario_consumption}
\end{equation}
where $\mathcal{I}_{\text{web}}$ denotes the web-based authoring interface,
$\mathcal{C}_{\text{local}}(\cdot)$ denotes a deterministic local consumption operator,
and $\mathcal{O}$ represents the structured outputs produced by automated analysis
pipelines.

Once scenarios are selected and materialized through the web interface, they transition from data-access abstractions into executable experimental units. In this stage, each canonical scenario $\mathcal{S}$ is treated as an input to local processing workflows, where additional structural transformations are applied to support controlled analysis and evaluation. In particular, scenario execution operates over temporally extended contexts derived from anchor instants selected during interactive exploration, enabling the systematic reconstruction of dynamic scene evolution around moments of interest.

\section{System Architecture} \label{architectureSec}
The system architecture defines how user-selected normalized multimodal data are transformed into executable scenario instances and systematically consumed by local analysis workflows. Rather than operating directly on full acquisitions, the platform adopts a scenario-based abstraction in which curated scenario contexts $\mathcal{S}$ are treated as atomic experimental units. These units are subsequently expanded in time and enriched with controlled contextual structure to support reproducible testing, model comparison, and behavioral evaluation.

\subsection{Temporal Scenario Extend}
\label{sec:scenario_generation}

Although scenario definitions are selected through the web-based interface, temporal scenario extension is executed locally over the canonical scenario contexts provided by the platform consumption layer. Each scenario $\mathcal{S}$ contains one or more normalized scene snapshots selected during interactive exploration and treated as semantic anchors for subsequent construction. A selected snapshot corresponding to a moment of interest is denoted as a temporal anchor $t_0$. For each temporal anchor, the platform generates a scenario
window by applying a configurable temporal expansion specified in seconds. In
particular, users define how many seconds of context are included before and after
the anchor instant by selecting the pre-event and post-event extents \(k\) and
\(m\), respectively. The resulting temporal window is defined as

\[
\Delta T = [\, t_0 - k,\; t_0 + m \,],
\]
where \(k\) and \(m\) denote the pre-event and post-event temporal extents,
respectively. This windowing strategy introduces controlled temporal context around
the anchor moment, enabling the inclusion of dynamic evolution immediately preceding
and following the event of interest. By explicitly defining temporal bounds during
scenario generation, all scenarios follow consistent and reproducible structural
rules, while allowing asymmetric temporal configurations when required by the
analysis task.

\begin{figure}[t!]
    \centering
    \includegraphics[width=0.55\columnwidth]{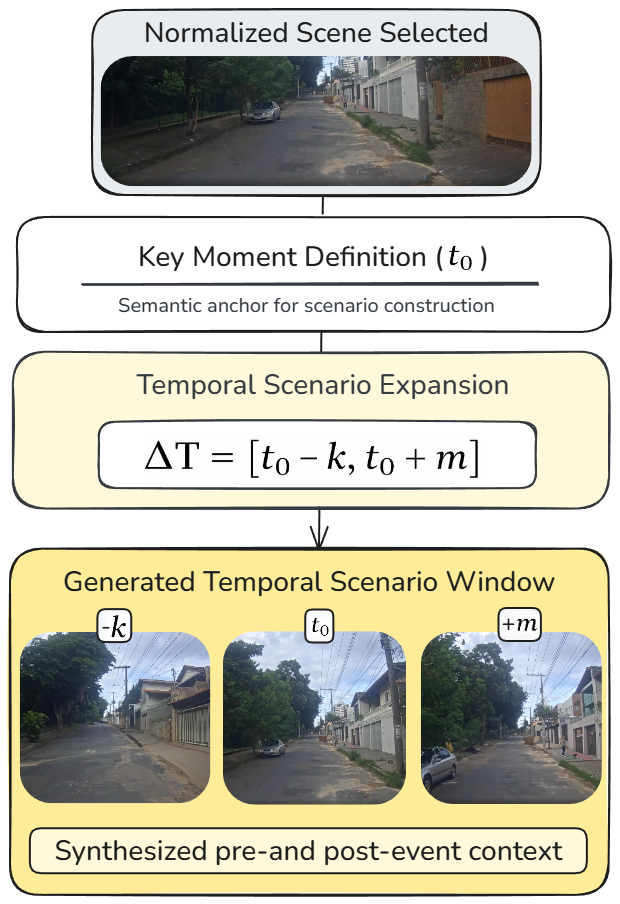}
    \caption{Scenario window generation from a single user-defined key moment
    (\(t_0\)), illustrating controlled temporal expansion before and after the
    anchor instant.}
    \label{fig:scenario_generation}
\end{figure}

The resulting scenario window is composed of a compact sequence of normalized
temporal units centered on the user-defined anchor moment and constrained to the
predefined temporal span. All multimodal information available in the normalized
data layer within this interval is included, allowing each scenario to capture a coherent and self-contained representation of the traffic situation without
introducing additional preprocessing or inference stages. The scenario generation procedure is fully deterministic, meaning that identical temporal configurations
always produce identical scenario instances, which is essential for ensuring
reproducibility across repeated analyses and experimental executions
\cite{gundersen2018reproducibility,pineau2021improving}. As illustrated in
Fig.~\ref{fig:scenario_generation}, this controlled temporal expansion establishes
each scenario as a stable and reusable analytical unit, forming the foundation for
subsequent interpretation and evaluation stages.

\subsection{Scenario Interpretation via Prompted LLM Evaluation}
\label{sec:prompt_engineering}

In addition to the scenario-generation stage, the platform incorporates an explicit prompt-engineering layer that primarily operates through the local consumption interface. This layer is responsible for transforming structured temporal scenarios into controlled analytical tasks suitable for language-model-based interpretation while maintaining strict alignment with the underlying normalized multimodal data.

Prompt construction operates over fixed temporal scenario windows. %The central concept underlying prompt construction is the use of a fixed temporal window associated with each generated scenario. Each scenario is centred on a user-defined key moment and expanded by a predefined number of seconds before and after this moment, forming a bounded temporal context. As a result, language model inputs are strictly limited to observations within the selected window, ensuring that interpretation is grounded exclusively in temporally consistent, observable data. In practice, scenarios are processed independently through organized batch workflows, in which each temporal window is treated as an isolated analytical unit and queued for evaluation, enabling systematic, scalable processing across large numbers of scenarios.

Based on this windowed representation, the platform supports different prompt configurations that define how each scenario is presented to the language model. These configurations include fixed prompt templates, designed to enforce standardized analytical tasks across all scenarios, and customizable prompt formulations, which allow alternative objectives or perspectives to be specified while preserving the same underlying scenario structure. In all cases, prompt content is systematically derived from the normalized data associated with the temporal window, including ego-vehicle dynamics, detected objects, road attributes, and environmental context. Decoupling scenario definition from prompting enables reuse across models and configurations. %By decoupling scenario definition from prompt formulation, the prompt engineering layer enables the reuse of identical scenario instances across multiple analytical configurations and evaluation settings. This design enables fair and reproducible comparisons between language models, while supporting flexibility in task definition. Prompt engineering is therefore treated as an explicit architectural component rather than an ad hoc procedure, reinforcing methodological rigor across all language model-based analyses conducted within the platform.

\subsection{Prompt Specification for Cognitive Risk}
\label{sec:prompt_global_constraints}

The platform adopts a structured prompt specification to constrain how language models represent and express traffic risk. Rather than producing free-form textual explanations, models must operate within a closed semantic space defined by numeric codes, ensuring consistency, comparability, and machine-readability across all evaluated scenarios and models.

The risk representation space encodes multiple complementary dimensions of traffic safety, including overall severity, conflict types, ego-vehicle behavior, vulnerable road user presence, temporal dynamics, uncertainty, and evidence attribution. Each dimension is expressed exclusively through predefined numeric identifiers, preventing semantic drift and subjective reinterpretation.
\begin{riskcodesbox}

\textbf{Overall Risk Severity} {\small\textit{(overall\_risk\_level)}}\\
{\small
0 Not identified;\;
1 Safe mode;\;
2 Low;\;
3 Moderate;\;
4 Elevated;\;
5 High;\;
6 Critical
}

\vspace{0.35em}
\textbf{Risk Occurrence Indicator} {\small\textit{(window\_has\_risk)}}\\
{\small
0 No risk identified (overall\_risk\_level = 0);\;
1 Risk identified (overall\_risk\_level = 1)
}

\vspace{0.35em}
\textbf{Evidence Attribution Signals} {\small\textit{(evidence\_signals)}}\\
{\small
1 Object presence;\;
2 Object distance;\;
3 Object--lane relation;\;
4 Speed;\;
5 Steering;\;
6 Braking;\;
7 Road/environment;\;
8 Image input
}

\vspace{0.35em}
\textbf{Risk Category Types} {\small\textit{(risk\_types)}}\\
{\small
2 Pedestrian;\;
3 Cyclist;\;
4 Rear-end;\;
5 Lateral conflict;\;
6 Intersection;\;
7 Speed;\;
8 Visibility;\;
9 Infrastructure;\;
10 Traffic density
}

\end{riskcodesbox}

Once the semantic space is fixed, a global prompt constraint layer is applied to ensure
deterministic behavior and strict epistemic discipline across all evaluated language
models. This layer defines the role of the model, limits admissible evidence sources,
and enforces a rigid output format, independently of any specific risk semantics.
Concretely, it operates as a fixed prompt prefix that constrains all responses to a
strictly structured JSON representation using only predefined numeric codes, thereby
transforming model outputs into machine-readable and directly comparable analytical
artifacts.

\begin{promptconstraintbox}

You are an expert in urban traffic risk assessment.

Using ONLY the information explicitly provided in the input
(structured CAN per second, YOLO detections including dist\_m and
lane\_rel, basic context, and IF PRESENT any provided images).

\medskip
\textbf{IMPORTANT:}
\begin{itemize}
\item Output MUST be valid JSON only. No markdown. No extra text.
\item Use ONLY the numeric codes provided (do not output strings).
\item Do NOT invent missing data.
\end{itemize}

\end{promptconstraintbox}

This structured schema transforms model outputs into directly comparable analytical artifacts. %By constraining both semantics and structure, this prompt layer transforms the language model output into a first-class analytical artifact, enabling systematic aggregation, comparison, and statistical evaluation across different models and scenario windows.

\begin{figure}[t!]
    \centering
    \includegraphics[width=0.55\columnwidth]{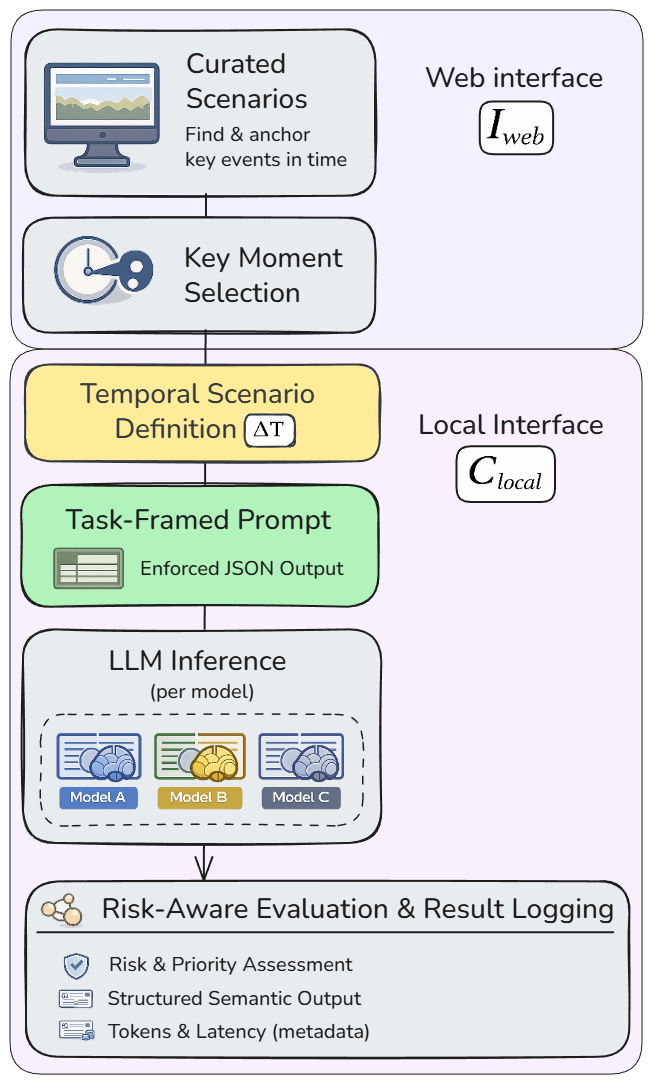}
    \caption{Scenario-driven workflow of the platform, from web-based identification of key moments to local temporal expansion and structured prompt construction. Curated scenarios are transformed into bounded temporal windows and independently evaluated by multiple language models under identical conditions, thereby establishing a deterministic, reproducible path from interactive scenario selection to controlled model inference.}
    \label{fig:scenario_generation}
\end{figure}

\subsection{Result Logging and Traceable Evaluation}
\label{sec:results_evaluation}

The final stage of the platform is dedicated to the systematic recording, aggregation, and analysis of results from language-model-based workflows. This stage consolidates automatically captured execution metrics with structured evaluation outputs, enabling a comprehensive assessment of model behavior under controlled, reproducible, and scalable experimental conditions. All results are persistently stored and explicitly linked to their corresponding scenarios, temporal windows, and prompt configurations, ensuring full traceability across evaluation runs and supporting longitudinal analysis.

Formally, the outcome of an evaluation run is represented as
\begin{equation}
\mathcal{R}
\;=\;
\mathcal{L}\!\left(\mathcal{S}_e,\;\mathcal{M}_e,\;\mathcal{P}_e\right),
\label{eq:results_logging}
\end{equation}
where $\mathcal{L}(\cdot)$ denotes the integrated logging and evaluation operator, and $\mathcal{S}_e$, $\mathcal{M}_e$, and $\mathcal{P}_e$ correspond to the sets of scenarios, language models, and prompt specifications effectively selected for a given experimental execution $e$. These sets are not assumed to be fixed or exhaustive; instead, they are defined according to the objective of each evaluation run.

This abstraction deliberately leaves the internal execution order and pairing strategy unspecified. Within this formulation, experimental designs may isolate individual factors, such as comparing different models under fixed scenarios and prompts, or evaluating prompt sensitivity for a single model, or jointly vary multiple dimensions or repeat identical configurations to assess consistency and robustness across executions. By decoupling the conceptual definition of results from implementation-specific scheduling details, the formulation preserves flexibility while maintaining strict traceability and comparability across overlapping evaluations.

\section{Experimental Setup} \label{experimentSec}
To evaluate the proposed dual-stage framework in interpreting complex driving environments, we conduct a series of experiments focused on "near-people" urban scenarios. This section details the experimental configuration, beginning with the protocol for selecting scenarios and curating data from the multimodal platform. We then describe the specific LLM configurations utilized as cognitive probes, the design of the deterministic prompt templates, and the structured evaluation metrics used to quantify model performance. By standardizing these experimental parameters, we establish a controlled environment to assess how effectively the framework handles semantic ambiguity and risk communication in safety-critical contexts.

\subsection{Collection and Evaluation Protocol}
\label{subsec:scenario_protocol}

\begin{figure}[t!]
    \centering
    \includegraphics[width=1\linewidth]{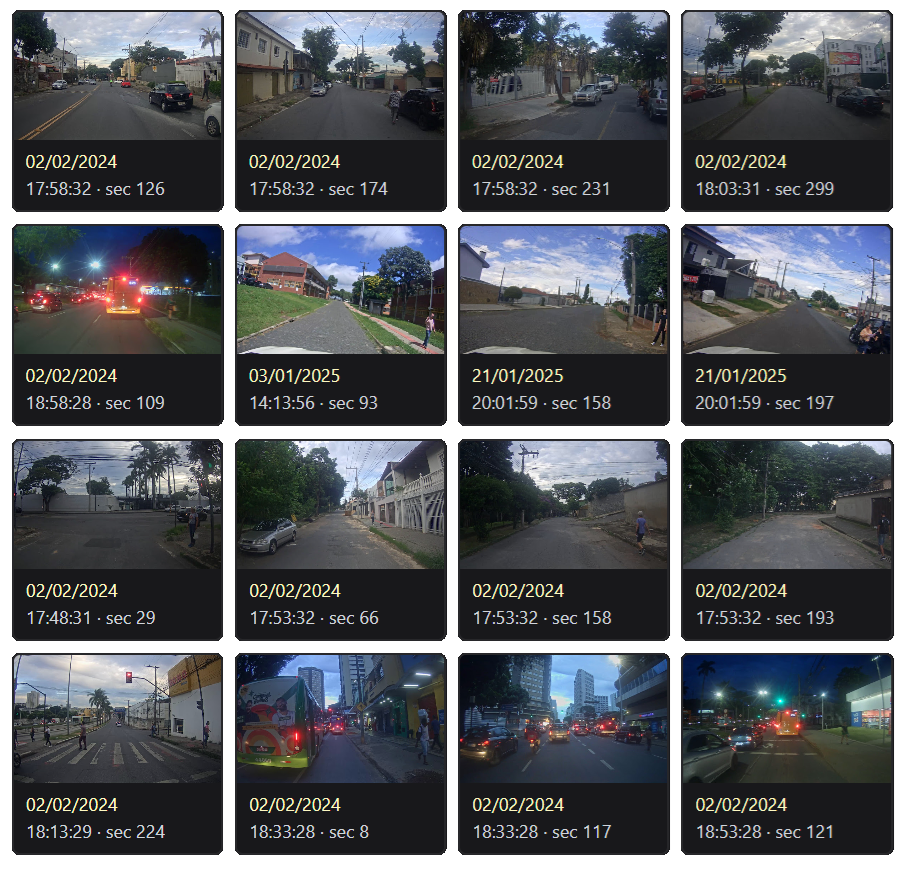}
    \caption{Mosaic of representative sixteen scene anchors, that serve as temporal seeds for subsequent scenario expansion, illustrating the diversity of near-people situations across urban layouts, traffic densities, lighting conditions, and pedestrian configurations.}
    \label{fig:scene_anchors}
\end{figure}

\begin{figure}[t!]
    \centering
    \includegraphics[width=0.6\columnwidth]{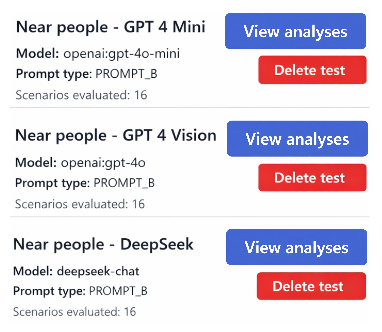}
\caption{Online visualization of the evaluated test sets available in the platform. Each entry corresponds to a \emph{near-people} key-moment collection evaluated under identical conditions by different language models, including two text-only configurations and one multimodal model.}

    \label{fig:near_people_ui}
\end{figure}

To ensure a controlled and semantically meaningful evaluation of language model behavior, all experiments were conducted over the same curated collection of key moments specifically designed to emphasize safety-critical urban interactions. Rather than operating over full-length acquisitions or arbitrarily sampled time windows, the evaluation focuses on temporally anchored scenario fragments centered around the presence of vulnerable road users in close proximity to the ego vehicle. In particular, a dedicated \emph{near-people} key-moment collection was constructed by identifying temporal anchors corresponding to situations in which pedestrians are spatially close to the ego vehicle. For each anchor time instant, a fixed temporal window spanning three seconds before and three seconds after the anchor was extracted, resulting in a standardized seven-second scenario window. This anchoring strategy follows the temporal expansion formulation introduced in the scenario generation framework, ensuring that each scenario captures both the contextual lead-up and the immediate evolution of the interaction. A total of sixteen distinct scenario windows were selected following this criterion and are available at \href{https://carcara.onrender.com/#/collections/69460424ef30b8bc66efc9dc/llm-tests/handler?testName=Near+people+-+GPT+4+Mini\&llmModel=openai\%3Agpt-4o-mini\&promptType=PROMPT_B}{carcara.com/69460/scenario-selected}. While larger-scale evaluations involving hundreds or thousands of scenarios are feasible within the platform, the present study deliberately adopts a small-scale experimental setting. This choice enables clearer interpretation of model behavior, facilitates detailed qualitative and quantitative analysis, and serves as an initial validation of the proposed evaluation pipeline before large-scale deployment.

Formally, the evaluation set is defined as a collection of $N=16$ temporally anchored
scenario windows, each centered on a key moment $t_0^{(i)}$ and symmetrically expanded
by three seconds before and after the anchor,
\begin{equation}
\Delta T^{(i)} = \left[\, t_0^{(i)} - 3,\; t_0^{(i)} + 3 \,\right], \qquad i = 1,\dots,16,
\end{equation}
yielding fixed-duration seven-second scenarios with the interaction event centrally
aligned.

All selected scenarios were evaluated under identical conditions by three different
language models: a lightweight text-only model, a higher-capacity text-only model,
and a multimodal model with visual grounding. Each model received the same structured
scenario representation and prompt specification, ensuring that observed behavioral
differences arise from intrinsic model characteristics rather than variations in
input data or experimental setup.

\subsection{Evaluated Language Models}

Scenario evaluations in this work adopt an organized batch-based execution strategy as a deliberate methodological choice, rather than as a fixed or inherent constraint of the platform. This strategy is selected to enable controlled, repeatable, and scalable comparison of language model behavior across heterogeneous driving scenarios, while preserving strict isolation between individual analytical contexts.

Both text-only and multimodal models are considered, enabling a systematic investigation of how access to visual information influences semantic scene interpretation, risk perception, and the generation of assistance-oriented recommendations. In the multimodal setting (GPT Vision), models receive raw visual inputs (images) jointly with structured, normalized textual descriptors, enabling internal encoding of visual cues such as object appearance, spatial relationships, and scene layout and their direct integration into the reasoning process. In contrast, text-only models (GPT 4-Mini and DeepSeek-Chat) operate exclusively on symbolic, structured textual representations derived from perception modules, relying on external scene descriptions rather than direct visual evidence.

\section{Results and Discussion} \label{resultsSec}
This section reports the empirical outcomes of a standardized, scenario-based evaluation that compares multiple language models under identical input and prompt constraints. For every model, the same prompt specification and the same curated set of near-people temporal windows are provided, ensuring identical inputs and output constraints across runs. %All analyses reported in this section are computed directly from the models’ structured JSON responses, which are automatically aggregated by the platform. %The evaluation targets four dimensions: overall risk severity, threshold-based high-risk escalation, breadth of evidence use, and dominant causal attribution, thereby enabling the identification of consistent model-level cognitive profiles under controlled conditions. 
An online view of the evaluated test sets %and the corresponding model-specific analyses 
is available at \href{https://carcara.onrender.com/#/collections/69460424ef30b8bc66efc9dc/llm-tests}{carcara.com/6946/llm-tests}.

\subsection{Model-Level Risk Interpretation: Analysis Logic and Derived Metrics}

Before presenting the quantitative comparisons, it is important to clarify the analytical logic used to interpret the structured outputs produced by the evaluated language models. Rather than treating the reported metrics as isolated statistics, the analysis follows a conditional reasoning scheme that maps numerical tendencies to qualitative behavioral profiles. This procedure enables a consistent interpretation of model behavior across identical scenario windows and prompt conditions.

\begin{AlgoBox}{1}{Conditional Semantic Analysis of LLM Risk Outputs}
\label{alg:llm_conditional_analysis}

\begin{algorithmic}[1]
\Require Parsed LLM outputs for a fixed scenario set, risk threshold $\tau$
\Ensure Model-level qualitative risk characterization

\State Group all scenario windows by language model $m$

\ForAll{model $m$}
\State $\mu_{\text{risk}} \gets$ mean overall risk level
\State $\rho_{\text{high}} \gets$ proportion of windows with risk $\geq \tau$ (threshold-based)
\State $\mu_{\text{evidence}} \gets$ mean number of distinct evidence signals
\State $\mathcal{F} \gets$ frequency distribution of dominant risk factors (primary category per window)

\If{$\mu_{\text{risk}}$ is high}
\State Mark model as \emph{risk-conservative}
\Else
\State Mark model as \emph{risk-tolerant}
\EndIf

\If{$\rho_{\text{high}}$ is high}
\State Indicate elevated sensitivity to critical situations
\EndIf

\If{$\mu_{\text{evidence}}$ is low}
\State Indicate narrow or underspecified reasoning
\Else
\State Indicate broader contextual grounding
\EndIf

\If{$\mathcal{F}$ is concentrated on few factors}
\State Indicate specialized risk perception
\Else
\State Indicate diversified risk awareness
\EndIf
\EndFor

\end{algorithmic}
\end{AlgoBox}

The conditional logic summarized in Algorithm~\ref{alg:llm_conditional_analysis} serves as an interpretative bridge between the raw structured outputs and the comparative analyses discussed in the following subsections. By explicitly defining how risk severity, threshold-based alert escalation, evidence usage, and dominant causal attribution are jointly interpreted, the analysis ensures that observed differences across models reflect distinct cognitive risk profiles rather than arbitrary metric fluctuations.

\subsection{Overall Risk Assessment Behavior}

\subsubsection{Mean Overall Risk Level}

The mean overall risk level captures the baseline risk posture adopted by each language model when interpreting identical near-people scenarios. This metric reflects how conservatively or permissively a model classifies safety-critical situations under fixed semantic and contextual constraints. The evaluated models exhibit clear stratification in their baseline risk interpretations. The multimodal model consistently assigns higher average risk levels across scenario windows, indicating a more precautionary stance. In contrast, the text-only models show lower mean risk values, suggesting a more restrained interpretation of comparable situations despite operating on the same structured evidence.

Although the number of evaluated scenario windows is limited, the observed ordering remains consistent across the dataset, indicating a stable baseline tendency rather than random variation. Notably, this divergence emerges even though all models operate under the same semantic schema and prompt specification, with the multimodal model differing solely through visual grounding rather than additional symbolic inputs.

\begin{figure}[t!]
    \centering
    \includegraphics[width=0.8\columnwidth]{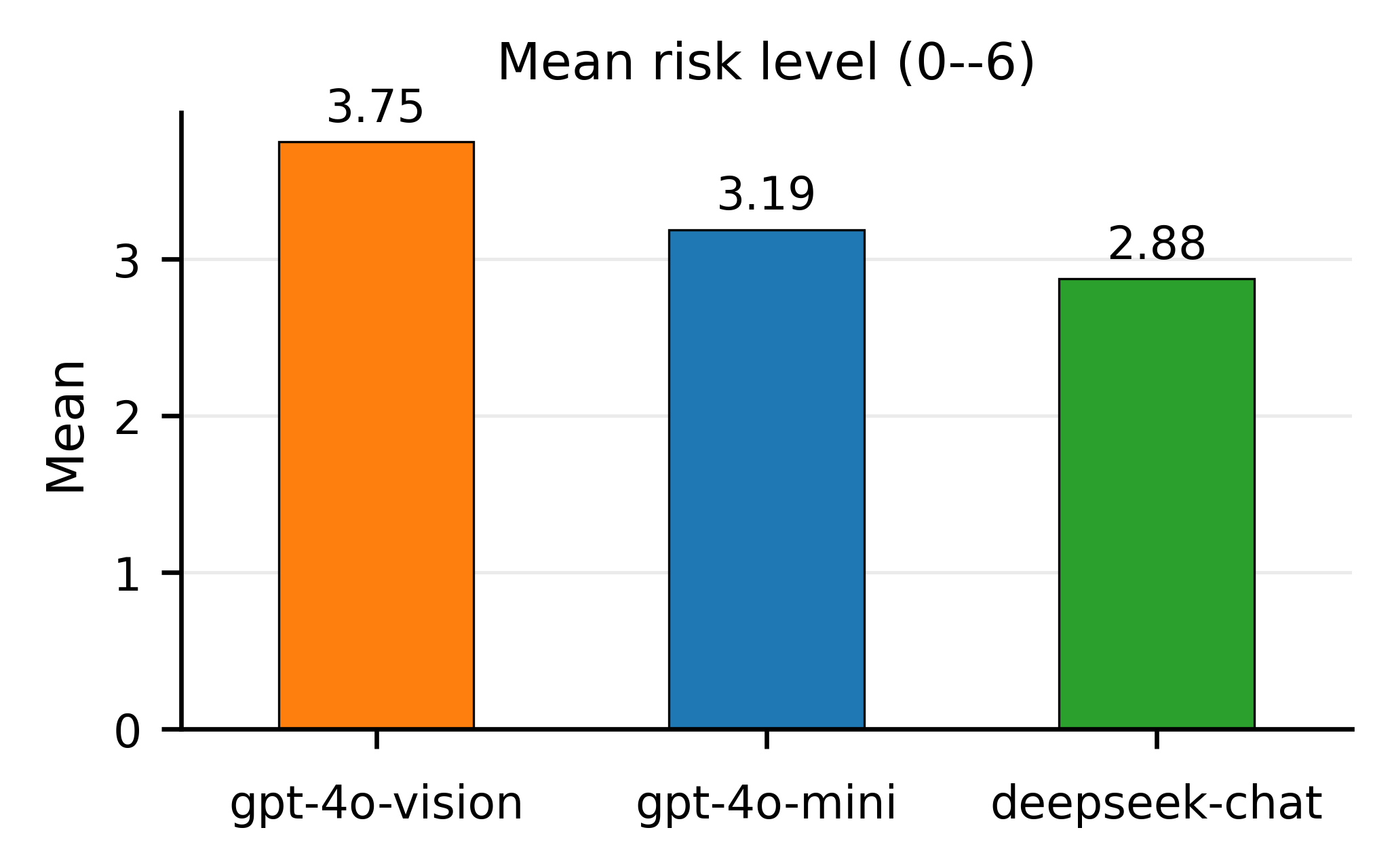}
    \caption{Mean overall risk level assigned by each evaluated language model across all scenario windows.}
    \label{fig:mean_risk}
\end{figure}

As illustrated in Fig.~\ref{fig:mean_risk}, these differences indicate that baseline risk interpretation varies systematically across models, even when scenario content and prompt structure are held constant. This initial posture establishes distinct starting points for subsequent reasoning stages, thereby influencing how each model approaches subsequent alert-escalation and causal-attribution analyses.

\subsubsection{High-Risk Escalation Tendencies}

Beyond baseline risk interpretation, the threshold-based proportion of scenario windows escalated to high risk provides insight into how often each model signals that situations require elevated attention. In this analysis, high risk is defined as scenario windows with an overall risk level of four or higher, indicating elevated or more severe safety conditions. This metric captures alert-escalation tendencies rather than the average risk posture, approximating how often a cognitive module would trigger strong driver-facing warnings under identical conditions.

The evaluated models exhibit clear differences in escalation behavior. The multimodal model escalates a larger fraction of scenario windows to high risk, indicating heightened sensitivity to potentially critical situations. In contrast, the text-only models adopt more selective escalation strategies, classifying fewer windows as high risk despite operating on the same structured evidence and prompt constraints. This pattern suggests that visual grounding is associated with both higher perceived severity and more frequent escalation of situations to alert-worthy states.

Importantly, escalation frequency and mean risk level reflect related but distinct dimensions of cognitive behavior. A model may maintain a comparatively elevated baseline risk assessment while selectively escalating, or, conversely, escalate frequently despite moderate average risk scores. The observed divergence indicates that alert escalation constitutes an independent behavioral axis, shaping how aggressively a model prioritizes safety-relevant events.

\begin{figure}[t!]
\centering
\includegraphics[width=0.8\columnwidth]{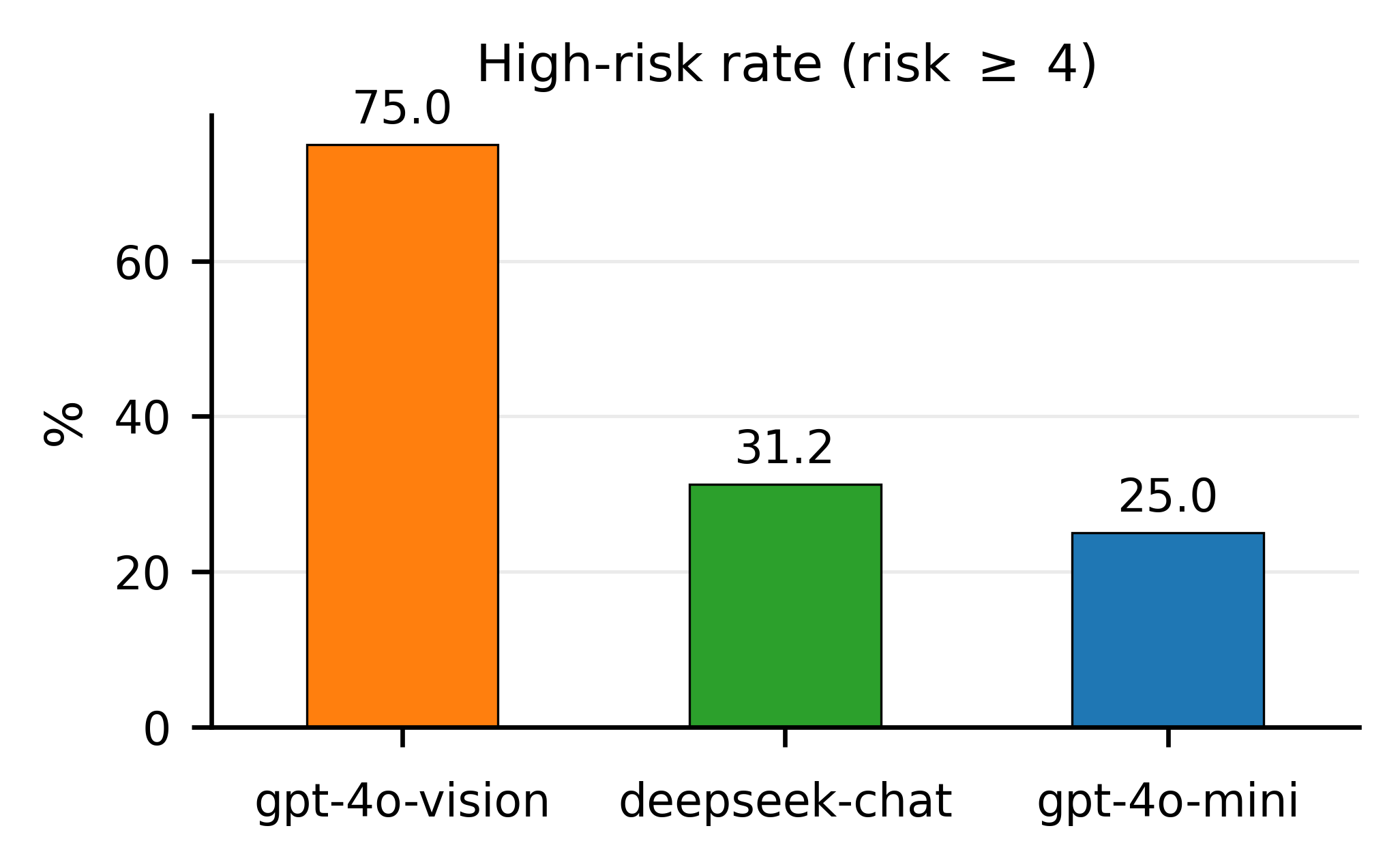}
\caption{Percentage of scenario windows classified as high risk (risk level $\geq 4$) by each model.}
\label{fig:high_risk_rate}
\end{figure}

From an ADAS design perspective, these differences suggest practical implications. Higher escalation rates may enhance early hazard salience in complex urban environments, but they also increase the risk of alert fatigue if not calibrated using context-aware thresholds or multi-stage alert policies. More conservative escalation strategies may reduce unnecessary alerts, but risk delaying the emphasis of marginal yet safety-relevant situations. These trade-offs are reflected in the distribution of high-risk classifications summarized in Fig.~\ref{fig:high_risk_rate}.

\subsubsection{Evidence Usage and Reasoning Breadth}

The mean number of distinct evidence signals explicitly referenced by each model provides a proxy for reasoning breadth and for externalizing justification. Operationally, this metric is computed as the number of unique evidence signals listed in the structured evidence priority field of each model output. In this evaluation, evidence signals correspond to structured contextual elements available in the scenario representation, such as ego-vehicle dynamics, object detections and distances, lane relations, road attributes, and environmental cues. Because the prompt enforces a constrained JSON schema, this metric reflects which signals the model chooses to externalize as explicit justification rather than the full set of information it may have processed internally.

The text-only models exhibit higher average evidence counts, indicating a more enumerative explanation style that verbalizes a broader set of supporting cues. In contrast, the multimodal model references fewer distinct signals, producing more compact justifications. This difference suggests that access to visual input may reduce the need for explicit textual enumeration of contextual cues, as some salient information can be internally grounded in the image. As a result, a trade-off emerges between conciseness and traceability: broader evidence externalization improves auditability, whereas more compact outputs favor brevity at the potential cost of transparency.

\begin{figure}[t!]
    \centering
    \includegraphics[width=0.8\columnwidth]{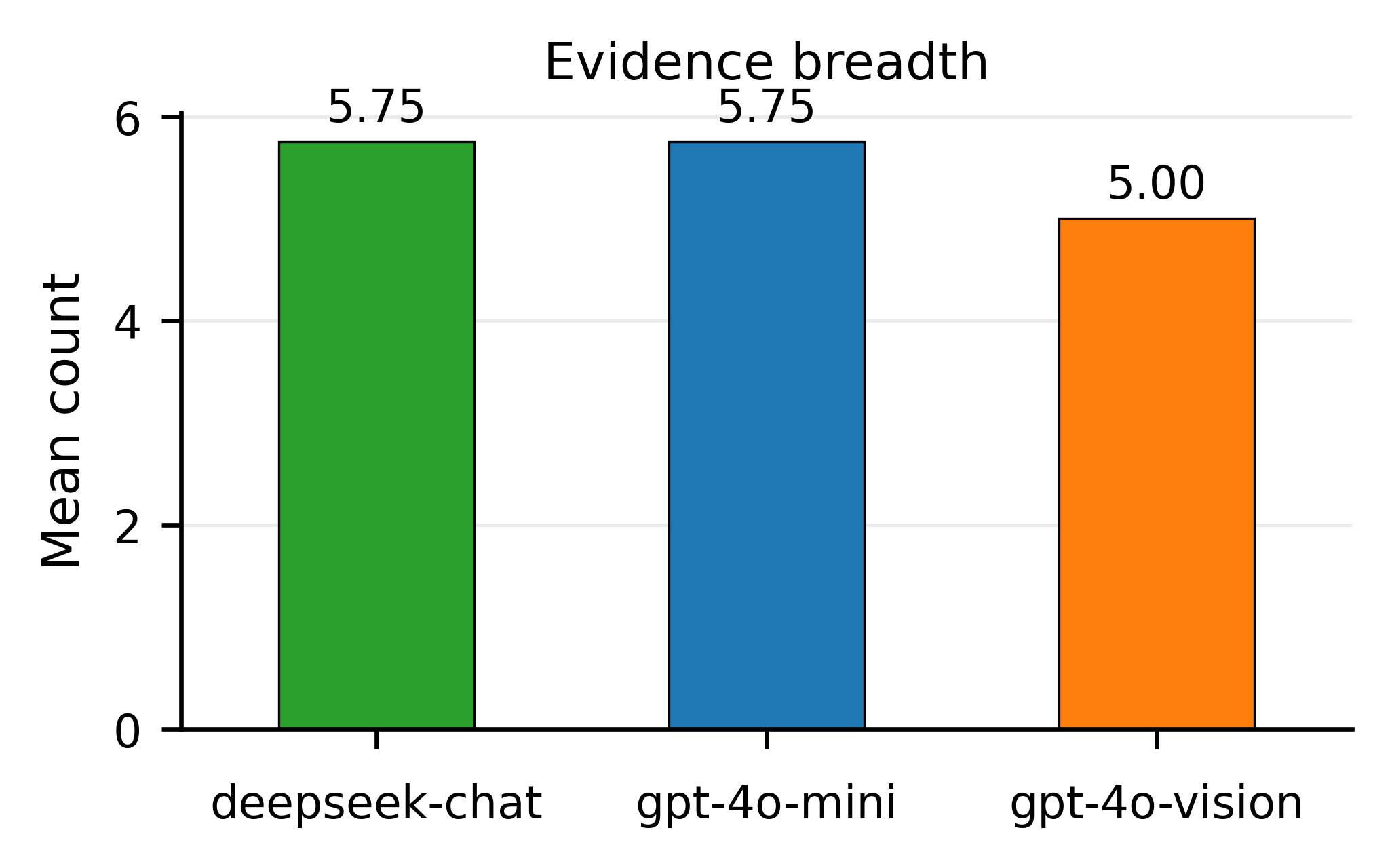}
    \caption{Mean number of distinct evidence signals explicitly referenced in model outputs, reflecting reasoning breadth and justification externalization.}
    \label{fig:evidence_count}
\end{figure}

From a methodological perspective, these findings emphasize that explainability in LLM-based ADAS extends beyond decision correctness to encompass an externalization policy for evidence. Multimodal grounding may enhance perceptual richness and risk sensitivity, while simultaneously reducing the explicit articulation of non-visual cues. For driver-facing assistance, additional prompting or post-processing may be required to ensure that critical signals, such as object distance, lane-relative position, and ego dynamics, are consistently reported even when visual context is available. These trends are quantitatively summarized in Fig.~\ref{fig:evidence_count}.

\subsection{Risk Factor Attribution Patterns}

\begin{figure}[t!]
    \centering
    \includegraphics[width=0.8\columnwidth]{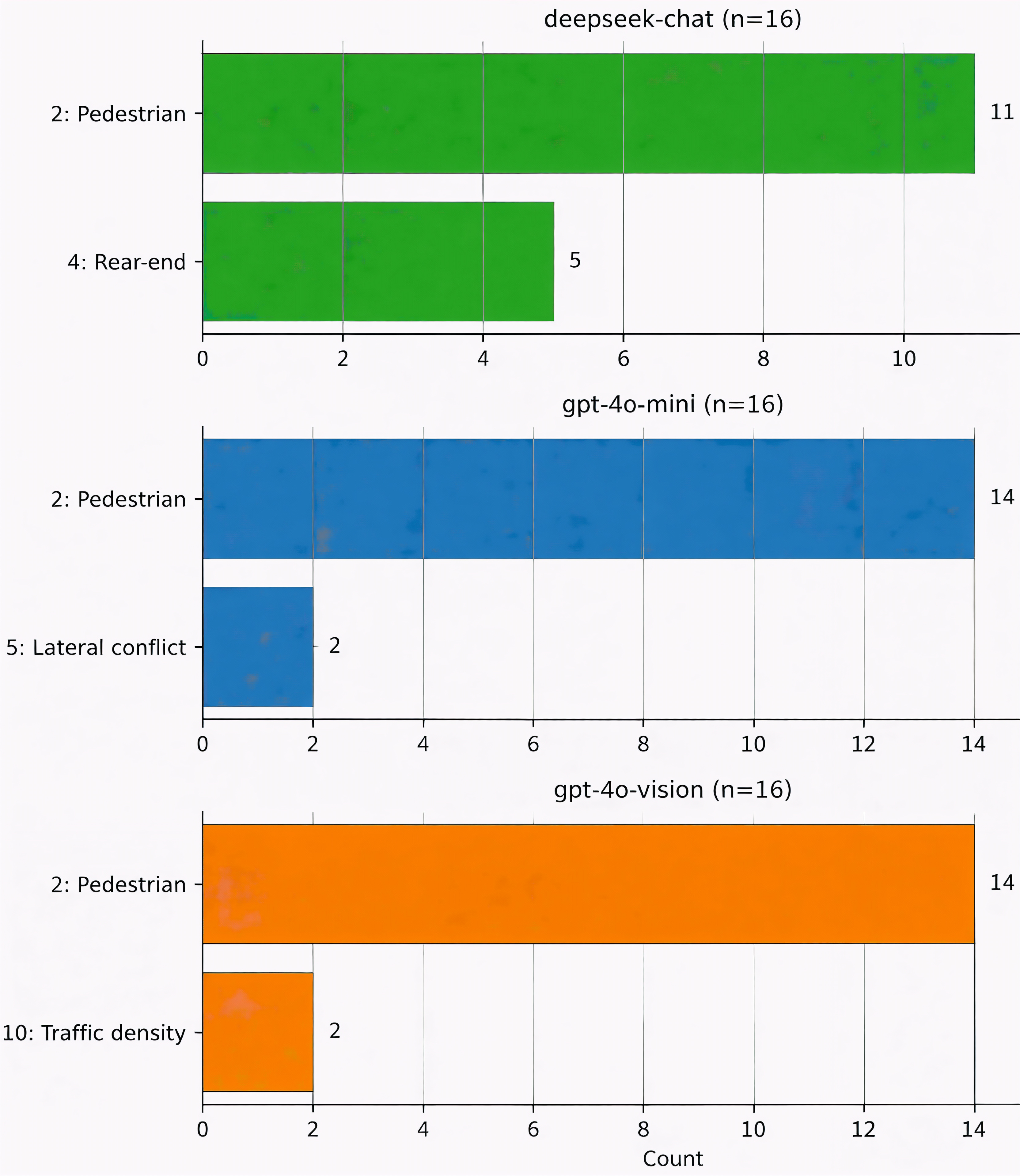}
    \caption{Distribution of dominant risk factors identified by each model. The six most frequent factors are shown explicitly, with the remaining factors grouped as Other.}
    \label{fig:risk_factors}
\end{figure}

Beyond magnitude and escalation, the attribution of risk provides insight into how models frame causal explanations under identical evidence. While all evaluated models operate on the same structured scenario inputs, they differ in how they assign primacy to specific risk sources, revealing distinct patterns of causal emphasis.

In this analysis, the dominant risk factor corresponds to the primary causal category explicitly selected by the model for each scenario window. Across all models, pedestrian-related risk emerges as the dominant attribution, which is expected given that the test set is intentionally curated around near-people interactions and reflects strong safety priors toward vulnerable road users. This shared primary attribution suggests a consistent baseline alignment with ADAS safety objectives when vulnerable road users are present.

However, once the primary factor is fixed, the attribution structure diverges substantially across models. Although only a single dominant factor is selected per scenario window, the relative frequency of non-primary categories reveals systematic differences in secondary attribution tendencies across models. The DeepSeek text-only model most frequently elevates rear-end conflicts as the next dominant category, suggesting a tendency toward interaction geometry and collision archetypes, inferred from object proximity and ego-vehicle dynamics. The GPT-4o-mini text-only model, in contrast, more often selects lateral conflict as its dominant non-pedestrian category, indicating a stronger emphasis on side interactions and spatial lane-relative relations. The multimodal GPT-4o-vision model exhibits a qualitatively different secondary attribution pattern, more frequently elevating traffic density as the prominent non-pedestrian factor, reflecting a broader contextual framing in which global scene complexity becomes salient once visual cues are available.
\begin{figure*}[t!]
    \centering
    \includegraphics[width=0.9\textwidth]{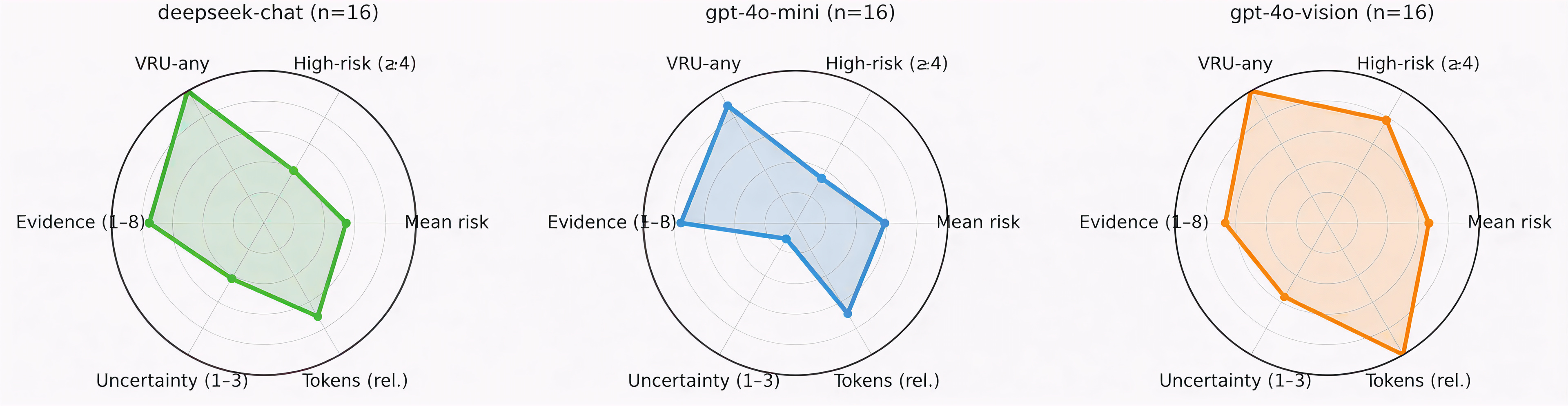}
    \caption{Radar-based summary of inter-model behavior across near-people scenarios, highlighting systematic differences in risk attribution, high-risk escalation, VRU presence interpretation, evidence signal count (reasoning breadth), uncertainty expression, and relative token usage. The visualization illustrates how identical scenario artifacts yield divergent yet internally consistent interpretations across text-only and multimodal LLM configurations.}
    \label{fig:radar}
\end{figure*}
These attributional differences are summarized quantitatively in Fig.~\ref{fig:risk_factors}, which reports the distribution of dominant risk factors selected by each model across all evaluated scenario windows. The figure makes explicit that, beyond the shared pedestrian-centric primary risk, each model systematically privileges different secondary explanations given the same evidence.

From a cognitive ADAS perspective, these observed differences are nontrivial. Even when models agree that a scenario is risky and converge on the presence of a vulnerable road user, they may guide driver attention toward different aspects of the environment. A model that foregrounds collision archetypes (rear-end or lateral conflict) may produce alerts that are more directly actionable for immediate defensive maneuvers, whereas a model that foregrounds traffic density may emphasize situational vigilance and scene-level caution. These findings indicate that model choice affects not only how much risk is communicated, but also how surrounding contextual risks are framed, foreshadowing deeper sources of inter-model ambiguity in VRU presence interpretation.

\subsection{Inter-Model Ambiguity and Composite Uncertainty Across Scenarios}
\label{subsec:intermodel_ambiguity}

The analyses presented in the previous subsections reveal consistent inter-model behavioral differences in baseline risk assessment, threshold-based escalation, evidence externalization, and dominant causal attribution, even under identical scenario inputs and fixed prompt constraints. While each of these dimensions provides an isolated view of model behavior, a more comprehensive picture emerges when these divergences are jointly considered. In this work, inter-model ambiguity is therefore treated as a composite phenomenon arising from the combined interaction of multiple semantic and decision-related dimensions, rather than from disagreement in any single metric.

To operationalize this notion of ambiguity, a two-stage formulation is adopted. First, ambiguity is detected using a minimal divergence criterion, in which any scenario window where at least one model deviates from inter-model consensus is flagged as ambiguous. Second, the degree of ambiguity is quantified using continuous disagreement measures jointly derived from the four previously analyzed dimensions: overall risk severity, high-risk escalation, evidence usage breadth, and dominant risk factor attribution. This formulation enables a scenario-centric characterization of uncertainty that captures not only whether models disagree, but also how strongly and consistently such disagreement manifests across multiple dimensions. The resulting composite uncertainty score is normalized to the $[0,1]$ interval, where $0$ denotes full inter-model agreement across all evaluated dimensions and models, and $1$ denotes maximal divergence, corresponding to consistent disagreement across all dimensions. While ambiguity detection itself follows a minimal criterion, the associated uncertainty magnitude reflects graded, structured disagreement rather than a purely binary outcome.

Across the evaluated test set of sixteen near-people scenarios, the resulting composite uncertainty scores reveal a structured distribution of inter-model alignment rather than uniform disagreement. Based on the magnitude of the composite uncertainty, the scenario set partitions naturally into three uncertainty tiers. The low-uncertainty subset, denoted as $\mathcal{S}_{\text{low}} = \{S_3, S_2, S_{14}, S_{15}, S_4\}$, corresponds to cases in which all evaluated models remain largely aligned across the considered dimensions, reflecting situations in which semantic interpretation and risk assessment are consistently resolved. The intermediate-uncertainty subset, $\mathcal{S}_{\text{med}} = \{S_5, S_9, S_{16}, S_1, S_8, S_{11}\}$, reflects partial divergence typically confined to one or two dimensions, such as differences in evidence prioritization or threshold-based escalation. Finally, the high-uncertainty subset, $\mathcal{S}_{\text{high}} = \{S_6, S_{12}, S_7, S_{13}, S_{10}\}$, concentrates scenarios exhibiting pronounced inter-model divergence across multiple dimensions simultaneously, representing structurally ambiguous situations in which multiple semantic and decision-related interpretations remain plausible. In aggregate, these subsets satisfy $(|\mathcal{S}_{\text{low}}| = 5,\ |\mathcal{S}_{\text{med}}| = 6,\ |\mathcal{S}_{\text{high}}| = 5)$, highlighting that inter-model ambiguity is graded and scenario-dependent rather than uniformly distributed.

\begin{figure}[t!]
    \centering
    \includegraphics[width=0.45\columnwidth]{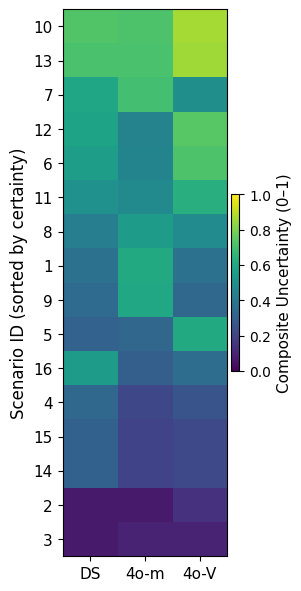}
    \caption{Compact heatmap of composite inter-model uncertainty across the evaluated scenarios. Rows correspond to scenario identifiers, ordered from lowest to highest uncertainty (bottom to top), and columns correspond to the evaluated language models. Color intensity indicates the per-model contribution to composite uncertainty, aggregated over overall risk severity, high-risk escalation, evidence usage, and dominant risk factor attribution.}
    \label{fig:intermodel_uncertainty}
\end{figure}

Figure~\ref{fig:intermodel_uncertainty} summarizes this analysis using a compact heatmap representation. Each row corresponds to a scenario, labeled by its original scenario identifier and ordered from lowest to highest composite uncertainty, while each column corresponds to one of the evaluated language models. Color intensity reflects the per-model contribution to composite uncertainty, aggregated across the four dimensions. By construction, the composite uncertainty score is normalized to the $[0,1]$ interval, where values close to $0$ indicate strong inter-model alignment across all dimensions, and values approaching $1$ indicate cumulative disagreement across multiple dimensions. This ordering enables a visual ``certainty-to-ambiguity'' trajectory, highlighting how inter-model divergence increases gradually across scenarios rather than appearing as isolated outliers.
From a methodological perspective, this composite uncertainty representation serves as an explicit audit layer that bridges individual metric comparisons and holistic scenario interpretation. Rather than treating inter-model disagreement as noise or isolated anomalies, the proposed formulation reveals structured, reproducible ambiguity patterns under controlled evaluation conditions. From an ADAS and SOTIF perspective, these findings emphasize that safety-relevant uncertainty arises not only from perceptual limitations but also from divergent semantic interpretations of partially specified scenarios. Explicitly modeling and visualizing such graded ambiguity is therefore essential for scenario-centric validation, model auditing, and the design of robust decision-support pipelines that remain resilient under semantic indeterminacy.

\subsection{Discussion: Implications for Scenario-Centric Auditing and SOTIF-Relevant Validation}
\label{sec:results_synthesis}

%The following subsection shifts from reporting empirical observations to interpreting their implications for LLM auditing and ADAS design.

The radar visualization in Fig.~\ref{fig:radar} provides a consolidated behavioral signature of the inter-model behavioral patterns previously discussed, aggregating the primary risk-related dimensions evaluated across the near-people scenario set. Rather than introducing new measurements, this figure synthesizes the previously presented outcomes, enabling direct comparison of model-level risk profiles across identical scenario windows and fixed prompt constraints. The summary highlights systematic divergence across models, most notably in baseline severity attribution and high-risk escalation frequency. Consistent with the earlier analysis, the multimodal configuration adopts a more precautionary stance, exhibiting a higher mean risk and more frequent escalations beyond the high-risk threshold, whereas the text-only models remain comparatively conservative despite operating on the same structured scenario artifacts. This synthesis supports the core methodological premise that controlled scenario reuse reveals stable, model-specific risk-interpretation profiles that are both comparable and reproducible. These divergences are consistent with SOTIF-style performance limitations, where hazardous outcomes can arise without malfunction when evidence is incomplete or semantically underdetermined.

Beyond the primary risk dimensions, the visualization further incorporates uncertainty expression (as encoded in the JSON output) and relative token usage as secondary behavioral indicators, offering deeper insight into how models operationalize justification and confidence. Taken together, text-only models more explicitly externalize uncertainty while producing longer outputs, reflecting a strategy that favors broader evidence enumeration and explicit justification. In contrast, the multimodal model exhibits lower relative token usage and more compact uncertainty signalling, suggesting a reliance on internalized perceptual grounding rather than extensive textual elaboration. This contrast reveals a trade-off between conciseness and audit traceability: while shorter responses may enhance efficiency and decisiveness, more verbose outputs can facilitate post-hoc inspection and accountability. In practice, this trade-off can be treated as a controllable design variable through prompt constraints and post-processing policies that enforce a minimum level of evidence disclosure for safety-critical alerts. Importantly, the joint analysis of risk severity, uncertainty, and token usage demonstrates that these dimensions are not independent; they jointly characterize distinct cognitive styles across models, supporting the role of LLMs as auditable cognitive probes in scenario-centric driving-assistance evaluation. Notably, these findings do not motivate using LLM outputs as ground-truth perception, but rather as auditable interpretive probes whose divergences can be measured, compared, and managed.

\section{Conclusion} \label{conclusion}

This study analyzed how LLMs interpret identical urban driving scenarios under structured, scenario-centric evaluation. Despite identical inputs and prompts, models exhibited systematic divergence in severity assessment, escalation behavior, evidence use, and causal attribution. Disagreement extended to the interpretation of vulnerable road users, reflecting intrinsic semantic indeterminacy rather than an isolated failure. These findings motivate explicit ambiguity modeling when integrating LLM-based reasoning into safety-aligned ADAS validation. Future work will extend this framework to large-scale scenario corpora and investigate how structured ambiguity metrics can inform adaptive alert policies and safety assurance processes under SOTIF-oriented validation.

%In this work, we analyzed how large language models interpret identical structured urban traffic scenarios within an ADAS-oriented reasoning framework. Using a controlled, scenario-based evaluation with constrained semantic outputs, the results showed that inter-model divergence arises systematically across multiple dimensions of risk reasoning, including baseline severity assessment, alert escalation behavior, evidence externalization, and causal attribution. These differences emerged despite identical prompts, inputs, and temporal context, indicating that they stem from intrinsic model-level semantic reasoning rather than experimental artifacts.

%A central finding is that this divergence extends to the interpretation of the presence of vulnerable road users (VRUs). Even in near-people scenarios, models produced distinct, internally consistent assessments of which agents were present, reflecting the semantic indeterminacy inherent to complex urban scenes under partial observability. From an ADAS perspective, this highlights a fundamental limitation of treating LLM-based reasoning modules as deterministic perception authorities. Instead, language models should be understood as interpretative components operating under uncertainty, motivating the explicit representation and management of semantic ambiguity when integrating LLMs into safety-critical decision-support pipelines.

\ifCLASSOPTIONcaptionsoff
  \newpage
\fi
 \bibliographystyle{IEEEtran} 
 \bibliography{main.bib}
 % \cite{evaluationLLMs,talk2drive,safetyLLMs,llmHumanFeedback}.
\end{document}